\documentclass[conference]{IEEEtran}
\usepackage{cite}
\usepackage{amsmath,amssymb,amsfonts}
\usepackage{graphicx}
\usepackage{textcomp}
\usepackage[table]{xcolor}
\usepackage{url}

\begin{document}

\title{A Reference Architecture for Integrating Agentic AI into Enterprise Software:
Responsibilities, Interactions, and Variability}

\author{%
\IEEEauthorblockN{Mingqin Yu}
\IEEEauthorblockA{UNSW Sydney\\Sydney, Australia\\mingchin.yuyu@gmail.com}
\and
\IEEEauthorblockN{Fethi Rabhi}
\IEEEauthorblockA{School of Computer Science and Engineering\\UNSW Sydney, Australia\\f.rabhi@unsw.edu.au}
\and
\IEEEauthorblockN{Liming Zhu}
\IEEEauthorblockA{CSIRO\\Sydney, Australia\\liming.zhu@csiro.au}
\and
\IEEEauthorblockN{Qinghua Lu}
\IEEEauthorblockA{CSIRO's Data61\\Sydney, Australia\\qinghua.lu@data61.csiro.au}
}

\maketitle

\begin{abstract}
Agentic AI moves generative AI applications toward systems that coordinate reasoning, enterprise context, model inference, and external actions. Integrating these capabilities into enterprise software introduces diverse architectural concerns spanning orchestration, knowledge and memory, model access, tools and actions, security and access control, runtime infrastructure, and evaluation. Existing research and industry architectures address many of these concerns, but differ in scope, abstraction, responsibility organization, interaction representation, and assumptions about which capabilities are required. This paper presents a technology-neutral reference architecture for enterprise agentic AI integration. We synthesize architectural capabilities from research publications, industry architecture guidance, and interoperability specifications according to architectural purpose. The architecture organizes these capabilities into ten responsibilities and defines boundary invariants, typed cross-responsibility interactions, and variation points that distinguish architectural responsibilities from condition-dependent capabilities and replaceable realization mechanisms. We instantiate the architecture as an ESG portfolio-analysis and pre-trade compliance agent built over an existing service-oriented ESG metric computation and reporting system, and complement the case with a mapping of an independently published agentic architecture. The case realizes seven of the ten responsibilities; different questions take different interaction paths, controls activate only under their conditions, and the probed boundary invariants hold. The independent architecture maps onto the proposed responsibilities without requiring an additional one. These results provide initial evidence for the architecture's organization and applicability rather than its completeness. The deployment and recorded results are available as a replication package.
\end{abstract}


\section{Introduction}
\label{sec:introduction}

Agentic AI extends generative AI applications from individual model responses toward goal-directed systems that can maintain context, reason about tasks, invoke external capabilities, and coordinate actions~\cite{abouali2025agentic}. In enterprise environments, these capabilities must operate within existing software, data, security, governance, and operational structures. An agentic system may need to access organizational knowledge, invoke one or more models, maintain execution state, interact with enterprise services, and operate under identity, authorization, safety, and operational controls. By integration we mean embedding agentic capabilities into this existing estate, including legacy systems, identity and access infrastructure, and governance, rather than building agents in isolation. Integrating agentic AI is therefore not only a model-selection or prompting problem, but also a software-architecture problem concerning how these heterogeneous capabilities are organized, connected, controlled, and operated.

Existing research and industry architectures provide substantial guidance for such systems. Enterprise-oriented work addresses concerns such as business alignment, workflow orchestration, organizational knowledge, governance, and enterprise integration~\cite{koo2026enterprise}, while industry architectures describe runtime concerns including model access, knowledge and memory, tools, security, state, observability, and multi-agent coordination~\cite{aws2026agentic,microsoft2026multiagent,google2026components}. Interoperability specifications further provide mechanisms for interactions with tools and contextual resources or between independent agents~\cite{mcp2026spec,a2a2026spec}. These approaches address many capabilities relevant to enterprise agentic AI, but differ in scope, abstraction, responsibility organization, interaction representation, and assumptions about which capabilities are required in a particular system.

These differences create three related architecture-design problems. First, heterogeneous capabilities such as orchestration, model inference, enterprise context, external action, and access control require clear responsibility boundaries, even when their implementations are colocated or provided by the same platform. Second, dependencies among these responsibilities require explicit interaction semantics without implying that every system follows the same runtime pipeline. Third, capabilities such as multi-agent collaboration, persistent memory, asynchronous execution, human approval, or execution isolation may apply only under particular workload, integration, or governance conditions. A reference architecture for enterprise agentic AI should therefore distinguish architectural responsibilities, their interactions, condition-dependent capabilities, and the replaceable mechanisms through which they are realized. We investigate the following research questions:
\begin{itemize}
    \item \textbf{RQ1:} What architectural capabilities, interactions, and decision conditions are identified for integrating agentic AI into enterprise software systems?
    \item \textbf{RQ2:} How can these capabilities be organized into technology-neutral responsibility boundaries with explicit interaction semantics and variability?
    \item \textbf{RQ3:} How can the reference architecture be used to build a concrete enterprise AI system?
    \item \textbf{RQ4:} To what extent does the reference architecture hold when evaluated against concrete enterprise agentic AI systems?
\end{itemize}

To address these questions, we synthesize architectural capabilities from research publications, industry architecture guidance, and interoperability specifications into ten responsibilities, grouped into design context (L0), interaction and access (L1--L2), agentic execution and enterprise integration (L3--L6), and runtime and operational support (L7--L9). We then define boundary invariants, typed cross-responsibility interactions, and variation points that make responsibility separation, interaction semantics, and condition-dependent choices explicit. We instantiate the architecture as an ESG portfolio-analysis and pre-trade compliance agent built over an existing service-oriented ESG metric computation and reporting system, and complement the case with a mapping of an independently published agentic architecture. The contribution is therefore not a new agent technology stack, but an architectural organization in which responsibility ownership remains stable while execution paths, optional capabilities, protocols, and technologies vary.

\section{Background and Related Work}
\label{sec:related-work}

\subsection{Agentic AI System Architecture}
\label{sec:agentic-architecture}

Agentic AI introduces multiple architectural concerns around model inference. Prior work characterizes agentic systems through capabilities including reasoning and planning, memory and state management, tool use, environmental interaction, and coordination of actions~\cite{abouali2025agentic,peykani2026holistic}. Broader reviews additionally identify areas such as networking, trust and safety, and evaluation~\cite{moralles2026slr}. These concerns are interdependent, and systems range from a single agent to multiple specialized agents that communicate, delegate, or collaborate~\cite{abouali2025agentic}.

Enterprise deployment broadens this problem: agentic execution must coexist with existing information, services, and controls~\cite{peykani2026holistic,moralles2026slr}.

\subsection{Reference Architectures for Enterprise Agentic AI}
\label{sec:enterprise-reference-architectures}

Existing architectures address enterprise agentic AI from different scopes and levels of abstraction. Koo and Chung take an enterprise-transformation perspective, organizing agentic AI around Business Alignment, Workflow Orchestration, Agent Collaboration, Organizational Knowledge, Governance Integration, and Enterprise System Integration~\cite{koo2026enterprise}. Industry architectures provide more implementation-oriented perspectives. AWS describes agents together with concerns including model access, tools, knowledge, security, observability, and discoverability~\cite{aws2026agentic}, while Microsoft's Multi-Agent Reference Architecture focuses on systems of collaborating agents and addresses orchestration, agent discovery, memory and state, knowledge access, external tools, security, and observability~\cite{microsoft2026multiagent}. Google Cloud similarly describes components and patterns for agentic applications across different agent, workload, and tenancy configurations~\cite{google2026components,google2026multitenant}.

Enterprise-oriented work treats governance, knowledge, and integration at a high level, whereas implementation-oriented architectures decompose them into runtime services or controls~\cite{koo2026enterprise,aws2026agentic,microsoft2026multiagent}, and application-specific architectures add domain-specific structure; DevCoach, for example, adds context-refinement loops, bounded tool use, validation, and audit artifacts for a software-engineering lifecycle~\cite{elsisi2026devcoach}. At the level of an individual agent, Lu et al.\ propose a pattern-oriented reference architecture for foundation-model-based agents covering interaction engineering, memory, planning, execution, and responsible-AI plugins~\cite{lu2024responsible}; it addresses agent design rather than enterprise integration. Interoperability specifications standardize specific interactions: MCP specifies how AI applications access externally exposed tools and contextual resources~\cite{mcp2026spec}, whereas A2A addresses interoperability between independent agents~\cite{a2a2026spec}. Such specifications complement architectural descriptions, but do not by themselves determine how the broader set of enterprise responsibilities should be organized.

Existing approaches also assume different configurations, from workload-specific patterns~\cite{google2026components} to multi-agent concerns that may not arise for a single agent~\cite{microsoft2026multiagent}; the related work is therefore substantial but fragmented. This motivates a technology-neutral synthesis in which responsibility boundaries, cross-responsibility interactions, and architectural variability are represented explicitly while implementation technologies remain replaceable realization choices.

\section{Research Method}
\label{sec:method}

\begin{figure*}[t]
    \centering
    \includegraphics[width=\textwidth]{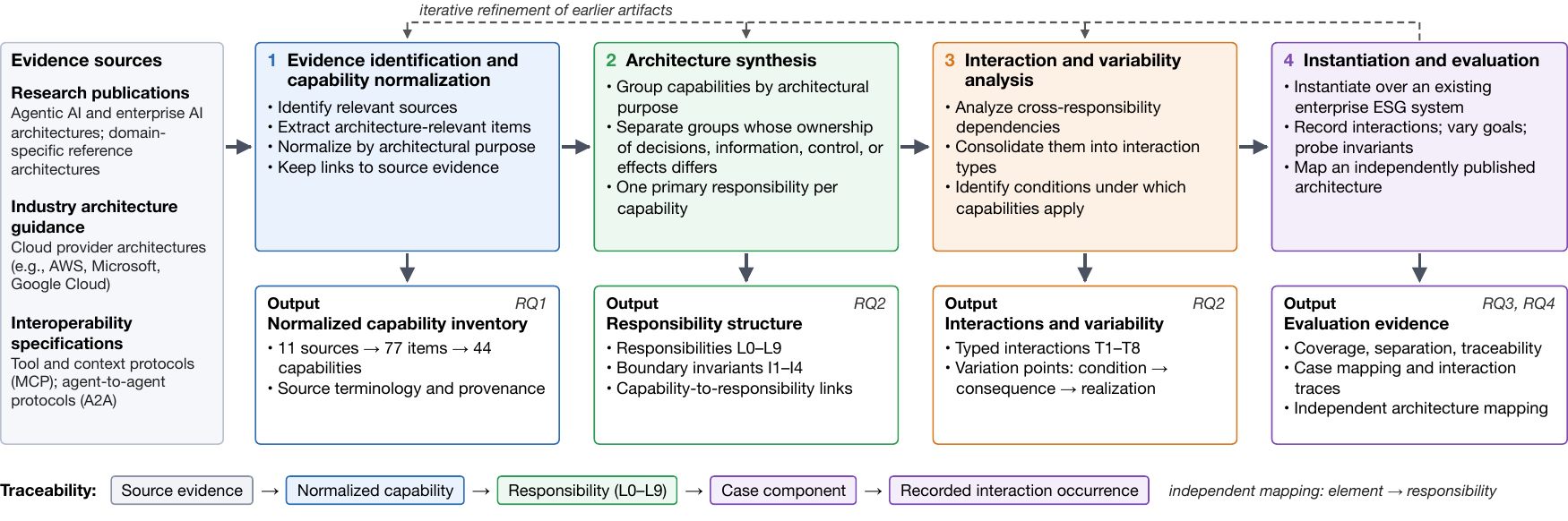}
    \caption{Research method for developing and evaluating the reference architecture, following empirically grounded reference-architecture design~\cite{galster2011empirically} and ProSA-RA~\cite{nakagawa2014consolidating}.}
    \label{fig:research-method}
\end{figure*}

We used an architecture-synthesis process comprising four activities (Fig.~\ref{fig:research-method}): (1) identifying architectural evidence and normalizing candidate capabilities, (2) synthesizing the normalized capabilities into technology-neutral architectural responsibilities, (3) analyzing cross-responsibility interactions and conditions of architectural variability, and (4) instantiating and evaluating the architecture. The process follows empirically grounded reference-architecture design~\cite{galster2011empirically} and the ProSA-RA progression from information-source investigation through architectural design to evaluation~\cite{nakagawa2014consolidating}. The activities correspond to the research questions: evidence (RQ1), synthesis (RQ2), instantiation (RQ3), and evaluation (RQ4). The activities were iterative during architecture development; the responsibilities, invariants, and interaction types were fixed before the evaluation reported in Section~\ref{sec:evaluation}.

\subsection{Evidence Identification and Capability Normalization}
\label{sec:evidence-normalization}

We conducted a targeted architecture-oriented review to identify capabilities, interactions, and decision conditions relevant to enterprise agentic AI integration. It covered three source categories: research publications describing agentic-AI and enterprise-AI architectures, industry architecture guidance describing deployable enterprise patterns and components, and interoperability specifications defining interactions among agents, contextual resources, and executable capabilities; following multivocal review guidelines, the latter two were treated as grey literature~\cite{garousi2019guidelines}. Sources were considered up to September 2026. Search terms combined agentic AI, AI agents, and enterprise AI with reference architecture, orchestration, tools, memory, security, observability, and interoperability. Sources were retained when they described architectural capabilities, relationships, controls, or configuration decisions relevant to integrating agentic AI into enterprise software; sources concerned only with model-level algorithms or application outcomes were outside scope.

For each retained source, we extracted architecture-relevant statements and recorded the corresponding capability, architectural purpose, relationships with other capabilities, applicable conditions where stated, and the source terminology. Because sources describe related concerns using different terminology or levels of abstraction, extracted items were normalized by architectural purpose rather than by product or technology name. Closely related items were consolidated when they served the same purpose, while distinctions were retained when they represented different responsibilities or decision conditions. The review retained 11 sources, from which 77 architecture-relevant items were extracted and consolidated into 44 normalized capabilities, each linked to the evidence from which it was derived; this inventory served as the evidence base for synthesis and for the analysis of interactions and variability.

\subsection{Architecture Synthesis}
\label{sec:architecture-synthesis}

We synthesized the normalized capabilities by architectural purpose rather than by source-specific terminology, product, or implementation technology. Capabilities were grouped when they contributed to the same primary responsibility, while capabilities with distinct purposes or control concerns were kept separate. Responsibility boundaries were refined by examining the ownership of architectural decisions and information across capability groups: whether a group primarily determines execution, provides contextual information, performs model inference, controls access, produces external effects, supports communication, operates infrastructure, or evaluates system behavior. A separate responsibility was retained wherever merging two capability groups would conflate distinct ownership of architectural decisions, information, control, or effects; where such groups nonetheless interact closely, the distinction was expressed as a boundary invariant.

Implementation mechanisms were kept separate from the responsibility structure. Technologies, protocols, storage mechanisms, deployment platforms, and architectural patterns were treated as possible realizations of capabilities rather than as responsibilities, so that a need such as contextual retrieval does not prescribe a particular database, protocol, model provider, or deployment technology. The structure was checked against the inventory and refined until each retained capability had exactly one primary responsibility.

\subsection{Interaction and Variability Analysis}
\label{sec:interaction-variability}

We then analyzed dependencies that cross responsibility boundaries. For each dependency, we examined the purpose of the interaction, the information or intent transferred, the direction of exchange, and which responsibility retained ownership of the associated decision or operation. Interactions with similar semantics were consolidated into interaction types, describing contracts between responsibilities without imposing a fixed end-to-end execution sequence.

We separately analyzed the conditions under which capabilities or interaction mechanisms apply. A capability was treated as a variation point~\cite{galster2014variability} when its inclusion or realization depended on characteristics such as agent topology, execution duration, integration requirements, external-effect risk, or operational constraints. For each variation point, we recorded the motivating condition, its architectural consequence, and representative realization mechanisms.

\subsection{Instantiation and Evaluation}
\label{sec:instantiation-evaluation}

We evaluated the architecture through two complementary analyses, following the case-based evaluation common for reference architectures~\cite{chand2026sra}. First, we implemented an instantiation as a deployed agent over an existing service-oriented ESG metric computation and reporting system and recorded its runtime interactions. We executed it against a representative portfolio question, portfolio and trade questions that differ in what they require, and probes of the boundary invariants. Second, we mapped an independently published agentic architecture that is not among the synthesis sources to the proposed responsibilities, to examine whether architectural elements defined outside the case can be represented using the same boundaries.

To answer RQ4, the evaluation considered three dimensions, one for each claim the architecture makes: that it can represent what systems require, that its boundaries are enforceable rather than nominal, and that its elements rest on evidence rather than on the case alone. \emph{Architectural coverage} examines whether the capabilities and elements required by each evaluated configuration can be represented by the proposed responsibilities and interactions. \emph{Responsibility separation} examines whether observed behavior is consistent with the ownership distinctions expressed by the boundary invariants. \emph{Traceability} examines whether architectural elements can be followed from source evidence and normalized capabilities to responsibilities and, for the concrete case, to implementation components and runtime interactions.

\section{Enterprise Agentic AI Integration Reference Architecture}
\label{sec:reference-architecture}

\begin{figure*}[t]
    \centering
    \includegraphics[width=\textwidth]{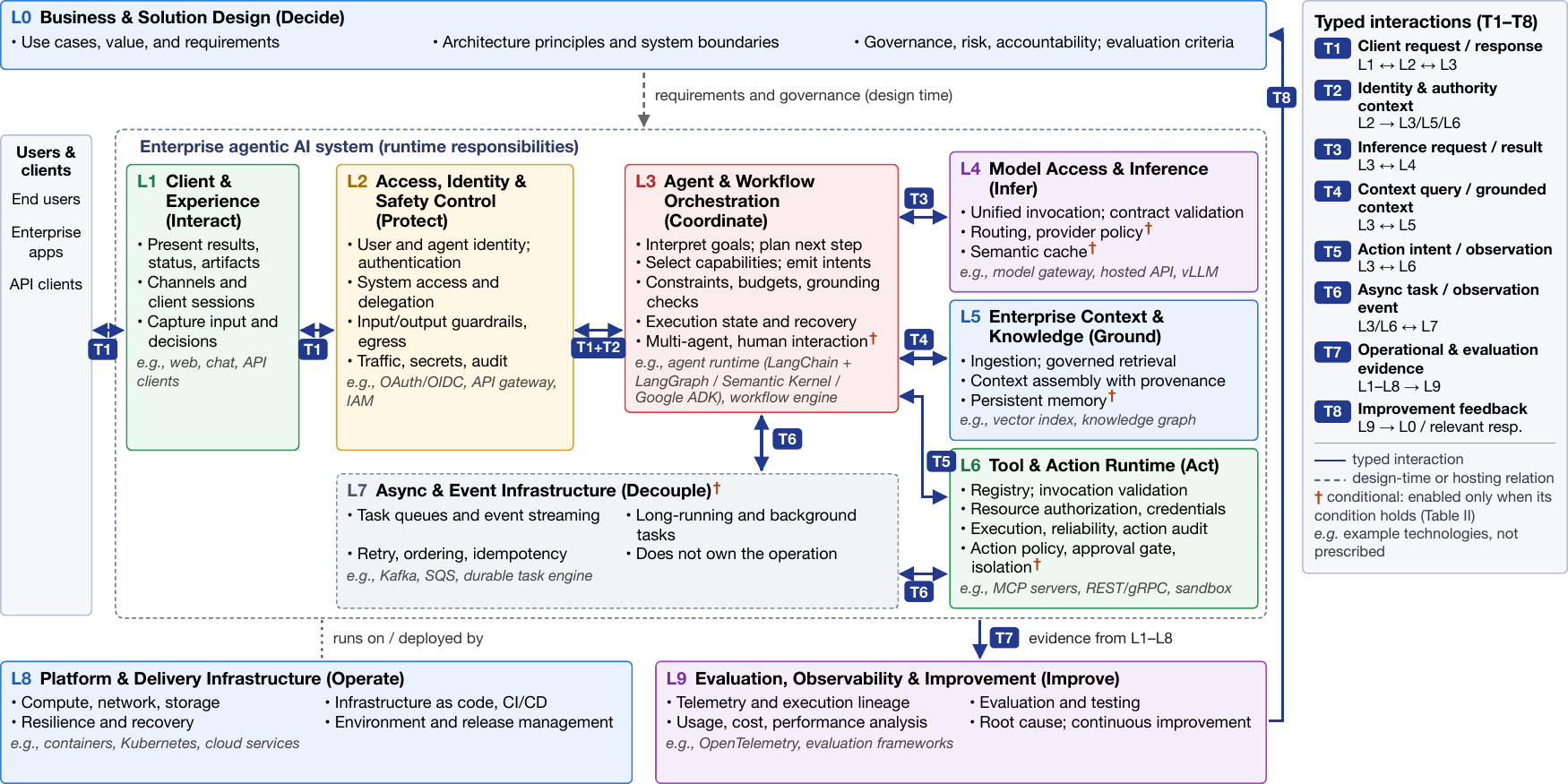}
    \caption{Enterprise Agentic AI Integration Reference Architecture. L0--L9 define architectural responsibility boundaries rather than a sequential processing pipeline; the example technologies are illustrative, not prescribed.}
    \label{fig:reference-architecture}
\end{figure*}

Addressing RQ1 and RQ2, the synthesis produced a reference architecture of ten responsibilities, denoted L0--L9. Each groups capabilities that share a primary architectural purpose and ownership concern: Business \& Solution Design (L0), Client \& Experience (L1), Access, Identity \& Safety Control (L2), Agent \& Workflow Orchestration (L3), Model Access \& Inference (L4), Enterprise Context \& Knowledge (L5), Tool \& Action Runtime (L6), Async \& Event Infrastructure (L7), Platform \& Delivery Infrastructure (L8), and Evaluation, Observability \& Improvement (L9). Fig.~\ref{fig:reference-architecture} presents this structure and its principal relationships.

L0--L9 are responsibility boundaries, not a processing pipeline: L0 sets design-time requirements, L8 hosts the runtime responsibilities, L9 receives their evidence, and during execution L3 selectively coordinates L4--L6, with L7 mediating asynchronous communication.

\subsection{Boundary Invariants}
\label{sec:architecture-overview}

The responsibility structure alone does not prevent closely related concerns from being merged or assigned ambiguously. We therefore define four boundary invariants: ownership distinctions that should hold across implementations. They constrain architectural responsibility rather than physical deployment: components implementing different responsibilities may be colocated, provided by the same platform, or deployed independently. The separation is logical; it becomes an enforceable security boundary only where credentials, privileges, and failure domains are also separated, which colocated components need not have.

\textbf{I1---Coordination--Execution Separation.} L3 owns execution coordination, including control flow and the decision to request an external capability, whereas L6 owns the authorization and execution of that capability against the target resource. L3 can therefore determine that an action should occur without assuming responsibility for producing the external side effect.

\textbf{I2---Execution-State--Knowledge Separation.} L3 owns information whose purpose and lifecycle are bound to coordinating an active execution, including execution state, intermediate results, and checkpoints. L5 owns information intended to be independently indexed or retrieved as reusable context across executions, including persistent enterprise knowledge and cross-execution memory.

\textbf{I3---System-Access--Resource-Authority Separation.} L2 establishes the principal, tenant, system-level access, delegated authority, and security context associated with a request. Responsibilities that own protected resources retain enforcement appropriate to those resources: L5 applies access and tenant constraints during retrieval, L6 enforces resource- and action-specific authorization and credential binding, and network reach is bounded by L2 egress control and L6 isolation. Requesting an action is distinct from being authorized to perform it: where execution is delegated to another agent or deferred, propagated authority should be scoped and time-limited, and L6 re-authorizes at execution time, because the originating context may have changed.

\textbf{I4---Evidence-Ownership--Correlation Separation.} Responsibilities L1--L8 remain authoritative producers of evidence about their local operations, while L9 correlates that evidence for end-to-end observation, evaluation, diagnosis, and improvement.

\subsection{Design Context, Interaction, and Access}
\label{sec:design-interaction-control}

\textbf{L0---Business \& Solution Design (Decide).} L0 translates business objectives and enterprise constraints into requirements for the agentic AI system. Its capabilities cover use cases and expected value, requirements, architectural principles and system boundaries, delivery and adoption, and governance across agents, models, prompts, knowledge, tools, and data. It also establishes ownership, accountability, risk, and regulatory requirements that constrain decisions elsewhere, rather than acting as a runtime step.

\textbf{L1---Client \& Experience (Interact).} L1 provides the interaction surfaces through which users and software clients access agentic capabilities, such as web applications, enterprise communication channels, and programmatic clients. L1 owns interaction and presentation, not agent reasoning or coordination.

\textbf{L2---Access, Identity \& Safety Control (Protect).} L2 protects the controlled system boundary and establishes the identity, authority, tenant, and policy context associated with a request. Its capabilities include ingress and traffic control, user and agent identity and authentication, system-level access authorization and delegation context, input and data protection, boundary-level safety controls, egress and data control, secrets management, and access auditing. L2 does not centralize all authorization or safety decisions: authority established at L2 is propagated to responsibilities that own protected resources, which retain enforcement specific to those resources (I3).

\subsection{Agent Execution and Enterprise Integration}
\label{sec:agent-execution-integration}

L3--L6 are most directly involved in goal-directed execution. Task allocation between model and conventional software is explicit: model-driven behavior is confined to selecting the next step (L3) and performing inference (L4), while access control, context retrieval, action execution, and every control are conventional software. Their separation allows reasoning, contextual grounding, and external effects to evolve independently while retaining explicit ownership at their boundaries.

\textbf{L3---Agent \& Workflow Orchestration (Coordinate).} L3 controls goal-directed execution. Its capabilities include agent runtime and control flow, execution constraints, in-flight execution context, prompt and runtime configuration, and execution state, recovery, and continuity; depending on requirements, it can additionally support collaboration among multiple agents and human interaction or intervention. It owns coordination decisions but not model inference, persistent enterprise knowledge, or execution of external side effects. A checkpoint can be durably stored without becoming enterprise knowledge; information meant to be reused across executions belongs to L5 (I2).

\textbf{L4---Model Access \& Inference (Infer).} L4 provides controlled access to model inference while isolating execution logic from particular model providers. Its capabilities include unified model access, model selection and routing, provider policy and capacity controls, model invocation, and inference-contract validation; semantic caching can be introduced when reuse is beneficial. L4 determines how an inference request is routed and executed, while L3 decides why inference is required and how its result affects execution.

\textbf{L5---Enterprise Context \& Knowledge (Ground).} L5 provides persistent enterprise information and governed context for agentic execution. Its capabilities span information sources, ingestion and preparation, enterprise context stores and persistent memory, retrieval and context assembly, and data governance. L5 owns the retrieval and assembly of persistent context, while L3 determines how that context is used; retrieval-augmented generation is therefore one possible realization of this relationship rather than the definition of the responsibility.

\textbf{L6---Tool \& Action Runtime (Act).} L6 resolves, validates, authorizes, executes, and records invocations of capabilities that interact with enterprise systems or produce effects outside agent coordination. Its capabilities include tool and capability registration, invocation validation and adaptation, resource-specific authorization and credential binding, tool integration and execution, reliability and outcome handling, and action audit and provenance. Where an external effect requires additional control, L6 can apply an action-policy or approval gate before execution. Under I1, L3 supplies the intent and execution context, and L6 determines whether and how the capability can be executed against the target resource and returns the observation, which L3 treats as untrusted input rather than instruction.

Guardrails and verifiers~\cite{lu2026safety} are not a separate responsibility; each is placed where the checked decision is owned. Input and output guardrails sit at the L2 boundary, retrieval constraints in L5, structural checks of model output in L4, and grounding and execution-constraint checks in L3. Pre-action verification of consequential effects sits in L6, whose action-policy or approval gate permits, refuses, or escalates an action, and L9 monitors from outside the execution loop. Treating retrieved context and tool results as untrusted input, together with the L3 grounding checks, addresses prompt injection through returned content.

\subsection{Runtime and Operational Support}
\label{sec:runtime-operational-support}

\textbf{L7---Async \& Event Infrastructure (Decouple).} L7 supports asynchronous, event-driven, and long-running communication through queues, event streaming, and delivery-reliability mechanisms such as retry, ordering, and idempotency. L7 is introduced when execution calls for temporal decoupling, background processing, durable delivery, or event-driven coordination, and changes delivery semantics without taking ownership of the operation communicated.

\textbf{L8---Platform \& Delivery Infrastructure (Operate).} L8 provides the compute, connectivity, storage, resilience, and delivery environment (infrastructure as code, CI/CD, release management) on which the other responsibilities operate. L8 owns platform operation rather than agent behavior: infrastructure autoscaling belongs to L8, whereas terminating a task that exceeds its budget belongs to L3.

\textbf{L9---Evaluation, Observability \& Improvement (Improve).} L9 correlates operational and evaluation evidence across the architecture to assess system behavior and support diagnosis and improvement. Its capabilities include telemetry, execution lineage, evaluation and testing, usage and cost analysis, root-cause analysis, and continuous improvement. Under I4, L1--L8 remain authoritative for events produced within their boundaries, while L9 links and analyzes this evidence across an execution.

\subsection{Typed Interactions and Variability}
\label{sec:interactions-variability}

Responsibility separation alone does not specify how architectural concerns cooperate during execution. We therefore characterize recurring cross-responsibility dependencies as typed interactions (Table~\ref{tab:interactions}). An interaction type identifies the architectural purpose of an exchange and the responsibilities involved without prescribing a transport protocol, deployment topology, or fixed execution sequence. MCP can realize selected interactions with contextual resources or executable capabilities, while A2A can realize communication with agents coordinated by L3 or, for agents owned by another party, reached through L6; REST, gRPC, or messaging provide alternative realizations.

\begin{table*}[t]
\centering
\caption{Typed Cross-Responsibility Interactions}
\label{tab:interactions}
\footnotesize
\setlength{\tabcolsep}{4pt}
\begin{tabular}{p{0.04\textwidth} p{0.20\textwidth} p{0.14\textwidth} p{0.54\textwidth}}
\hline
\textbf{Type} & \textbf{Interaction} & \textbf{Responsibilities} & \textbf{Architectural semantics} \\
\hline
T1 & Client Request / Response & L1 $\leftrightarrow$ L2 $\leftrightarrow$ L3 & Carries client intent through the controlled boundary at L2 to L3 and returns responses, status, or artifacts through L2 egress control, separating interaction concerns from execution coordination. \\
T2 & Identity \& Authority Context & L2 $\rightarrow$ L3/L5/L6 & Propagates principal, tenant, entitlement, delegated-authority, policy, and trace context; receiving responsibilities retain resource-specific enforcement. \\
T3 & Inference Request / Result & L3 $\leftrightarrow$ L4 & Requests model capabilities under specified constraints and returns model output, validation status, model metadata, and usage information. \\
T4 & Context Query / Grounded Context & L3 $\leftrightarrow$ L5 & Expresses an information need and returns governed context, evidence, metadata, and provenance appropriate to the execution. \\
T5 & Action Intent / Observation & L3 $\leftrightarrow$ L6 & Requests execution of an external capability and returns its result, status, error, state-change evidence, or provenance. \\
T6 & Async Task / Observation Event & L3/L6 $\leftrightarrow$ L7 & Mediates tasks and observations through asynchronous or event-driven delivery while preserving ownership in the originating and receiving responsibilities. \\
T7 & Operational \& Evaluation Evidence & L1--L8 $\rightarrow$ L9 & Provides locally produced traces, events, outcomes, usage, cost, audit information, and provenance for correlation and evaluation. \\
T8 & Improvement Feedback & L9 $\rightarrow$ L0 / relevant responsibilities & Returns evaluation findings, failure evidence, and quality, cost, or performance observations to inform governed architectural or configuration changes. \\
\hline
\end{tabular}
\end{table*}

We represent a variation point through the condition that motivates an architectural choice, the resulting architectural consequence, and representative mechanisms through which that consequence can be realized (Table~\ref{tab:variation}). Variability can therefore alter which capabilities are enabled and how responsibilities interact without redefining the responsibilities themselves. Synchronous and asynchronous configurations preserve the same L3 coordination and L6 action ownership while differing in whether L7 mediates their interaction. A single-agent configuration omits L3 collaboration capabilities without removing the L3 responsibility. To apply the architecture, an architect identifies the responsibilities a target system requires, enables condition-dependent capabilities according to its workload, integration, and governance requirements, selects realization mechanisms for the required interactions, and places controls according to the invariants. The responsibility boundaries and invariants remain stable while these choices vary.

\begin{table*}[t]
\centering
\caption{Representative Architectural Variation Points}
\label{tab:variation}
\footnotesize
\setlength{\tabcolsep}{4pt}
\begin{tabular}{p{0.28\textwidth} p{0.38\textwidth} p{0.26\textwidth}}
\hline
\textbf{Condition} & \textbf{Architectural consequence} & \textbf{Representative realization} \\
\hline
Multiple collaborating agents & Enable L3 agent collaboration, discovery, and coordination interactions & Agent registry; A2A or direct calls \\
Deterministic execution constraints & Constrain allowable execution transitions in L3 & State machine, DAG, or workflow rules \\
Human participation required & Enable L3 human interaction and intervention & Clarification, review, pause/resume \\
Protected or consequential external effect & Enable an L6 action-policy or approval gate & Policy service or human approval \\
Untrusted or high-risk execution & Isolate execution of the relevant L6 capability & Sandbox or restricted runtime \\
Delegated or deferred execution & Scope and expire propagated authority (T2); L6 re-authorizes at execution time & Scoped, short-lived tokens; re-authorization \\
Multi-turn or resumable sessions & Durably store L3 execution state; cross-session reuse belongs to L5 (I2) & Checkpointer or session store \\
Persistent cross-execution context required & Enable L5 persistent memory and governed retrieval & Vector, graph, document, or structured store \\
Multiple models or provider constraints & Enable L4 model routing and provider policy & Model gateway or router \\
Semantic reuse beneficial & Enable L4 semantic caching & Semantic cache \\
Long-running or background execution & Mediate relevant interactions through L7 & Queue or durable task mechanism \\
Event-driven coordination required & Enable L7 event infrastructure & Publish/subscribe or event stream \\
Recovery from multi-step state-changing operations & Coordinate recovery in L3 and compensating actions through L6 & Saga or compensation pattern \\
Standardized tool or context interoperability desired & Use standardized realization of selected L5/L6 interactions & MCP \\
Direct synchronous interaction sufficient & Bypass L7 for the relevant interaction & Direct API, RPC, or equivalent call \\
\hline
\end{tabular}
\end{table*}

\section{Reference Architecture Instantiation}
\label{sec:instantiation}

Addressing RQ3, we instantiate the reference architecture as an agent for portfolio carbon analysis and pre-trade checking against an illustrative ESG fund mandate, built over an existing ontology-driven ESG metric computation and reporting system~\cite{yu2026ontometric}. The case was selected because it involves concerns central to the architecture: enterprise knowledge retrieval, deterministic computation, model-driven coordination, role-based authorization, policy-controlled external action, and condition-dependent approval. The portfolio metric is the weighted average carbon intensity (WACI), the weight-averaged ratio of each holding's Scope~1 and Scope~2 emissions to its revenue~\cite{tcfd2021implementing}. The existing system computes company-level metrics through services and represents metric definitions in a knowledge graph, queried through competency questions (CQ1--CQ5) that return, for example, a category's metrics and how a metric is calculated. Section~\ref{sec:case-mapping} derives the agent's design by applying the steps of Section~\ref{sec:interactions-variability}; Sections~\ref{sec:case-example} and~\ref{sec:goal-variation} run it.

\subsection{Applying the Architecture}
\label{sec:case-mapping}

\emph{Required responsibilities.} The use case, the mandate limits, and the rule that a second person approves overrides are L0 decisions that set the requirements. At runtime, the agent must accept questions and approvals (L1), authenticate users and assign entitlements (L2), coordinate the steps of a question (L3), obtain model inference (L4), retrieve metric definitions (L5), and compute metrics, analyze portfolios, and act on trades (L6), all on a deployment platform (L8). Table~\ref{tab:case-mapping} lists the components. The knowledge graph (L5) and metric computation (L6) are reused from the existing system. A portfolio and compliance service added to L6 computes WACI and checks a trade deterministically against the fund's mandate, an illustrative mandate limiting portfolio intensity, exposure to high-intensity holdings, and single-holding concentration. The service holds the mandate; the request cannot change it. L3 runs a tool-use loop in which a language model chooses each next step among the capabilities of L5 and L6, or answers. L4 validates the inference contract of every model call: a response must call a registered tool or answer; refusals and truncated outputs count as failures. Only this step selection and inference are model-driven; all other components are conventional software.

\emph{Conditional capabilities.} Submitting a trade is a consequential external effect, so we enabled the L6 action policy and approval gate (Table~\ref{tab:variation}). On submission, L6 re-runs the mandate check and clears a compliant trade. For a breach with a requested override, it holds the trade until a principal who holds the override entitlement and is not the submitter approves it. No order management system is connected, so a cleared trade is recorded rather than executed. The other conditional capabilities were not enabled because their conditions are absent: requests complete synchronously in under a minute (no L7), one agent suffices (no L3 collaboration), the invoked services are trusted first-party services (no L6 isolation), and each question is answered independently (no L5 persistent memory). An L9 service for cross-execution correlation lies outside the evaluated deployment.

\emph{Realization mechanisms.} L5 and L6 are exposed through MCP~\cite{mcp2026spec}, the realization in Table~\ref{tab:variation} for standardized tool and context interoperability; L4 uses a hosted language model; all components run as separate containers.

\emph{Control placement.} The invariants determined where each control resides. Under I1, the model selects capabilities but does not execute them: the holdings and intensities passed to the portfolio and compliance service are assembled by L3 from L6 results, never supplied by the model, and L6, not L3, decides whether a trade is cleared. Under I3, L2 assigns entitlements by role (a portfolio manager may compute metrics, analyze portfolios, and submit trades; a compliance officer may approve overrides), and L6 enforces them at each invocation. Under I2, L3 retains only execution state, and definitions are retrieved from L5 for each request. Under I4, each responsibility records its own exchanges, carrying a trace identifier assigned at L2. L3 also bounds execution: the model may act only on the request's fund, holdings, and years; the number of model calls is bounded; and every number in the final answer must occur in the request or in a result returned by L5 or L6. A draft answer that fails this \emph{number check} is returned to the model once for revision.

\begin{table}[t]
\centering
\caption{Case components and the responsibilities they realize; * existing ESG system.}
\label{tab:case-mapping}
\footnotesize
\setlength{\tabcolsep}{4pt}
\begin{tabular}{l p{0.80\columnwidth}}
\hline
\textbf{Resp.} & \textbf{Case component} \\
\hline
L1 & Web service and portfolio analyst page \\
L2 & Access gate: signed session, role-based entitlements (portfolio manager, compliance officer), request validation, rate limit, egress check \\
L3 & Agent runtime: model-driven tool-use loop with execution constraints and number check \\
L4 & Model access: hosted language model, inference-contract validation \\
L5 & ESG knowledge graph* (competency questions), via MCP \\
L6 & Metric-computation service*, via MCP \\
L6 & Portfolio and compliance service (WACI, pre-trade mandate check), via MCP; entitlement checks; trade submission and override gate \\
L8 & Container deployment of all components \\
\hline
\end{tabular}
\end{table}

\subsection{Worked Example}
\label{sec:case-example}

\begin{figure*}[t]
    \centering
    \includegraphics[width=0.75\textwidth]{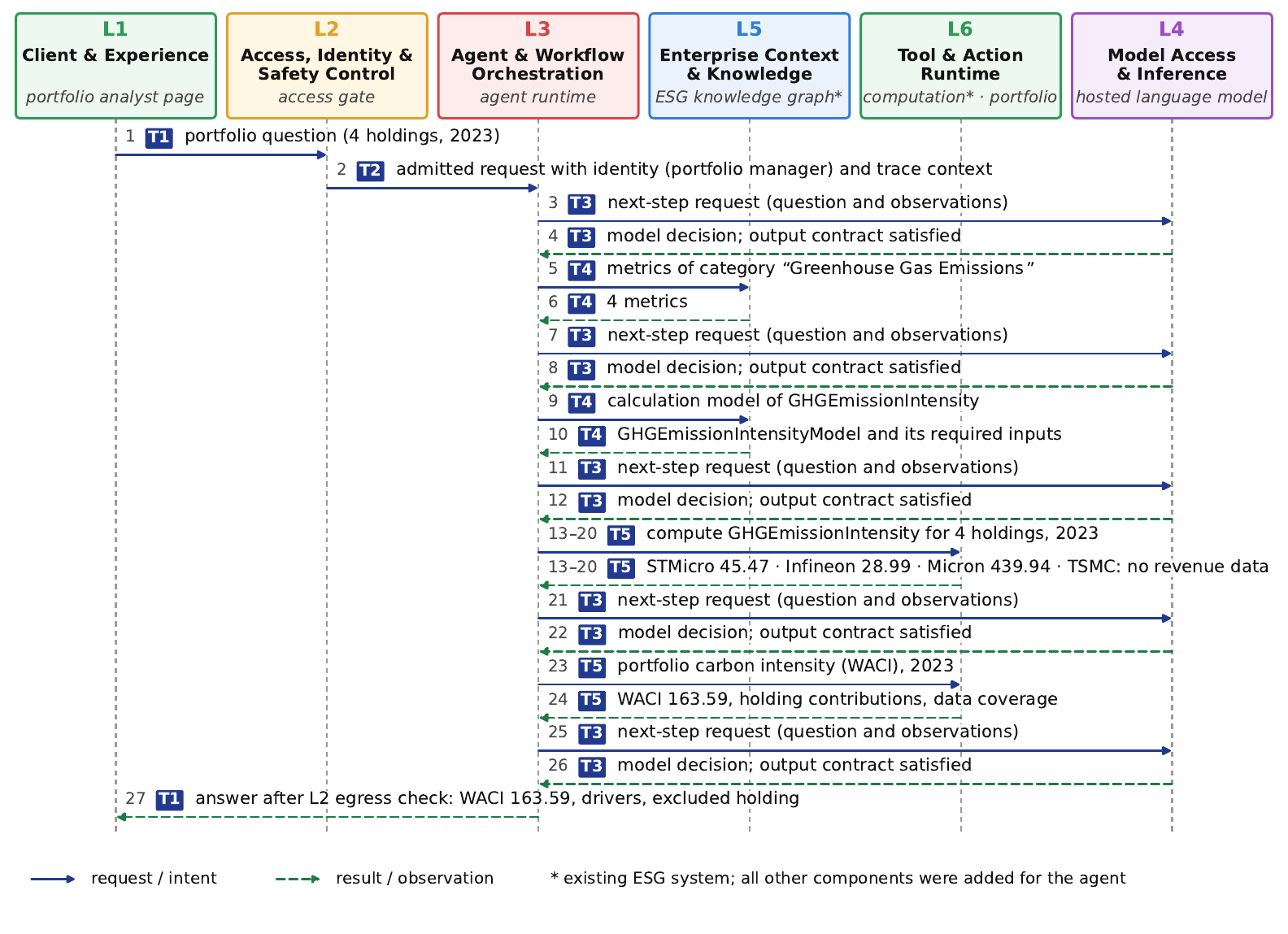}
    \caption{Recorded execution of the worked example. Each lifeline is a responsibility and the case component realizing it; components marked * belong to the existing ESG system. Each message is labeled with its occurrence number, which the text refers to, its interaction type from Table~\ref{tab:interactions}, and the content exchanged; the four per-holding computations (13--20) are drawn as one request and one result.}
    \label{fig:case-runtime}
\end{figure*}

In question G3 (Table~\ref{tab:goal-variation}), a portfolio manager holds four semiconductor companies (STMicroelectronics 30\%, Infineon 25\%, Micron 25\%, TSMC 20\%) and asks: \emph{What is the carbon intensity of my portfolio in 2023, and which holdings drive it?} Fig.~\ref{fig:case-runtime} shows the execution of this design as recorded. We count an \emph{interaction occurrence} as one directed exchange conforming to an interaction type of Table~\ref{tab:interactions}, a request and its return separately; numbers in parentheses below refer to the numbered occurrences in Fig.~\ref{fig:case-runtime}.

The request is admitted at L2 and passed to L3 with identity, role, and trace context (1--2). The model asks which metrics the \emph{Greenhouse Gas Emissions} category contains (5--6) and retrieves the definition of \emph{GHG Emission Intensity}, a calculation model $(\mathit{Scope\,1} + \mathit{Scope\,2}) / \mathit{Revenue}$ (9--10). It then requests the metric for each holding (13--20). The computation service returns 45.47, 28.99, and 439.94~t~CO$_2$e per USD million for STMicroelectronics, Infineon, and Micron, and reports that TSMC's revenue is unavailable. The model requests the portfolio figure (23--24): WACI, computed over the 80\% of the portfolio with data and re-weighted to it, is 163.59, of which Micron contributes 84\%. The answer, stating this figure, the drivers, and the excluded holding with its reason, passed the number check and returned to L1 (27). The execution comprises 27 occurrences realizing five of the eight interaction types, T1--T5.

\subsection{Goal-Conditioned Variation and Conditional Capabilities}
\label{sec:goal-variation}

To examine whether the responsibilities act as boundaries rather than as stages of a fixed pipeline, and whether the conditional capabilities activate only when their conditions hold, we executed the same deployment against seven portfolio and trade questions that differ in what they require (Table~\ref{tab:goal-variation}). Because the model's choices are not deterministic, each question was executed three times.

The seven questions produced goal-dependent execution paths with 5 to 41 interaction occurrences. Comparing years (G2) or adding holdings (G4) multiplied the computations. A trade within the mandate (G5) activated L6 policy enforcement and was cleared, while a trade breaching two limits (G6) additionally activated the approval gate. A question about share prices (G7), which the data cannot answer, was declined after a single model decision without engaging L5 or L6. Questions that only read data bypassed both controls. These conditions altered which capabilities were invoked and which controls became active, while responsibility ownership remained unchanged. Within a question, executions differed only in that one G5 execution also computed the current WACI and that four drafts (one for G2, three for G4) with model-derived figures were returned once by the number check.

\begin{table}[t]
\centering
\caption{Goal-conditioned variation over seven questions (three executions each; Occ.\ ranges reflect the variation described in the text).}
\label{tab:goal-variation}
\footnotesize
\setlength{\tabcolsep}{4pt}
\begin{tabular}{@{}l@{\hspace{4pt}}p{0.40\columnwidth}@{\hspace{4pt}}c@{\hspace{4pt}}c@{\hspace{4pt}}l@{}}
\hline
& \textbf{Question} & \textbf{Occ.} & \textbf{L6 calls} & \textbf{Outcome} \\
\hline
G1 & Intensity and drivers (4 holdings, all with data) & 27 & 5 & Answered \\
G2 & Change since 2022 & 39--41 & 10 & Answered \\
G3 & As G1, one holding lacking revenue (worked example) & 27 & 5 & Answered; gap reported \\
G4 & Intensity and drivers (7 holdings) & 35 & 8 & Answered \\
G5 & Trade within the mandate; submit & 31--33 & 6--7 & Cleared \\
G6 & Trade breaching the mandate; override & 33 & 7 & Held for override \\
G7 & Share prices & 5 & 0 & Declined \\
\hline
\end{tabular}
\end{table}

\section{Evaluation}
\label{sec:evaluation}

Addressing RQ4, we evaluate the instantiation of Section~\ref{sec:instantiation} and a mapping of an independently published architecture along the three dimensions of Section~\ref{sec:method}. Each checks a part of the design: whether the required responsibilities fit within L0--L9 (coverage), whether the controls placed under the invariants hold when probed (separation), and whether design decisions can be followed from source evidence to runtime behavior (traceability). Responsibilities are thus assessed through coverage and separation, interactions through the recorded occurrences, and variation points through goal-conditioned variation (Section~\ref{sec:goal-variation}).

\subsection{Architectural Coverage}

Case components were assigned to seven of the ten responsibilities (Table~\ref{tab:case-mapping}). Each had one primary responsibility, and none required a responsibility outside L0--L9. The seven instantiated responsibilities contain 33 of the 44 normalized capabilities of Section~\ref{sec:method}; this count is descriptive rather than a completeness score, because capabilities are condition-dependent. The executions realized five of the eight interaction types (T1--T5), and the two enabled conditional capabilities activated only when their conditions held (Section~\ref{sec:goal-variation}). The independent mapping (Section~\ref{sec:independent-mapping}) touches L0--L6 and L9. Across the two analyses nine of the ten responsibilities are represented; L7 appears in neither, because neither configuration is asynchronous.

\subsection{Responsibility Separation}

Each boundary invariant of Section~\ref{sec:architecture-overview} was probed with operations that would expose a violation if its boundary were not enforced (Table~\ref{tab:invariants}). All probe operations produced outcomes consistent with the invariants. For I4 the case shows evidence ownership and a shared correlation identifier but has no L9 correlation service, so I4 is assessed as consistent rather than preserved.

\begin{table*}[t]
\centering
\caption{Boundary-invariant probes and observed outcomes.}
\label{tab:invariants}
\footnotesize
\setlength{\tabcolsep}{4pt}
\begin{tabular}{p{0.13\textwidth} p{0.37\textwidth} p{0.35\textwidth} p{0.075\textwidth}}
\hline
\textbf{Invariant} & \textbf{Probe} & \textbf{Observed outcome} & \textbf{Result} \\
\hline
I1 Coordination--Execution & Submit question G1; count invocations of the L6 computation and portfolio services; compare the reported intensities with L6 results; inspect L3 for an execution mechanism. & One computation per holding and one portfolio calculation, each invoked through L6; reported intensities equal the L6 results; L3 exposes no execution mechanism. & Preserved \\
I2 Execution-State--Knowledge & Inspect the state L3 retains after the request; submit a second request and count knowledge-graph retrievals. & L3 retained only execution-local state (question, plan, computations, results, status); definitions and provenance came from L5 through CQ1--CQ5, and the second request retrieved them again. & Preserved \\
I3 System-Access--Resource-Authority & Admit a principal at L2 without the computation entitlement and invoke L6; submit a request with a forged session token; submit a trade breaching the mandate and approve an override as the portfolio manager, as the submitter holding the override entitlement, and as the compliance officer. & The admitted principal was refused at L6; the forged request was refused at L2 before coordination; the breaching trade was held, the override was refused for the portfolio manager and for the submitter, and approved for the compliance officer. & Preserved \\
I4 Evidence-Ownership--Correlation & Inspect the recorded occurrences for their emitting responsibility and correlation identifier. & Occurrences were emitted by four responsibilities, and the identifier established at L2 was carried into the L5 requests. & Consistent \\
\hline
\end{tabular}
\end{table*}

\subsection{Independent Architecture Mapping}
\label{sec:independent-mapping}

\begin{table*}[!t]
\centering
\caption{Mapping of an independently published agentic architecture~\cite{lu2024responsible} to the proposed responsibilities.}
\label{tab:independent-mapping}
\footnotesize
\setlength{\tabcolsep}{4pt}
\begin{tabular}{p{0.15\textwidth} p{0.41\textwidth} p{0.065\textwidth} p{0.31\textwidth}}
\hline
\textbf{Source element} & \textbf{Source-described purpose} & \textbf{Resp.} & \textbf{Observation} \\
\hline
Context engineering & Dialogue interface; goal creators interpreting the user's goals & L1/L3 & Spanning: interaction L1, goal interpretation L3 \\
Prompt/response engineering & Prompt and response generation from templates and personas & L3 & Prompt and runtime configuration \\
Memory & Short-term (in context window) and long-term memory & L3/L5 & Spanning: consistent with I2 \\
Planning & Plan generation and reflection, including human reflection & L3 & Coordination \\
Cooperation & Voting-, role-, and debate-based cooperation among agents & L3 & Conditional; the case uses one agent \\
Execution engine & Task executor using tools or agents; task monitor & L3/L6 & Spanning: consistent with I1 \\
Tool/agent selector & Find or generate tools and agents via a registry & L3/L6 & Spanning: agent discovery L3, tools L6 \\
Guardrails & Input, output, RAG, and execution guardrails & L2/L5/L6 & Spanning: consistent with I3 \\
Recording, risk assessment & Black box recorder, continuous risk assessor, explainer & L9 & Accountability and operational evidence \\
AIBOM, co-versioning & Provenance of tools, agents, models; model variants & L0/L4 & Spanning: provenance L0, model variants L4 \\
AI models & External, fine-tuned, or sovereign foundation models & L4 & Model access and selection \\
\hline
\end{tabular}
\end{table*}

The mapping serves two dimensions: whether an architecture designed without reference to the proposed one can be represented within L0--L9 (coverage), and whether its elements that cross the proposed boundaries divide where the invariants predict (separation). We mapped Lu et al.'s reference architecture for foundation-model-based agents~\cite{lu2024responsible}, which is not among the synthesis sources. Its mapped elements are the component groups of its architecture figure, with responsible-AI plugins grouped by concern. Elements were interpreted by their stated purpose, and those crossing a boundary were recorded as spanning. All eleven elements correspond to responsibilities within L0--L9, and none required an additional responsibility: four map directly, one is a conditional capability the case does not instantiate, and six span a boundary (Table~\ref{tab:independent-mapping}).

The spanning elements are the informative cases, because the source organizes components by agent function and the proposed structure by ownership. Its memory separates in-context working state from long-term knowledge retrieved into it, consistent with I2. Its execution engine both decides and performs tasks, which the proposed structure separates into coordination in L3 and capability execution in L6, consistent with I1. Its guardrails act at the user boundary, during retrieval, and around tool execution, consistent with resource-specific enforcement under I3. L7 and L8 have no counterpart, as the source addresses agent design rather than deployment. These distinctions arise independently of the case and the synthesis, but do not show that the boundaries are universal.

\subsection{Traceability}

For the case, each architectural element can be followed along the chain source evidence $\rightarrow$ normalized capability $\rightarrow$ responsibility $\rightarrow$ case component (Table~\ref{tab:case-mapping}) $\rightarrow$ recorded interaction occurrence (Fig.~\ref{fig:case-runtime}); the independent mapping adds the complementary link from external architecture element to responsibility (Table~\ref{tab:independent-mapping}). The deployment, experiment scripts, recorded results, and the generator for Fig.~\ref{fig:case-runtime} are available as a replication package.\footnote{\url{https://github.com/Inspiring-Ming/AI-Agent-for-ESG}}

\section{Discussion}
\label{sec:discussion}

\subsection{Architectural Implications and Trade-offs}

Separating orchestration from model access, context, and action lets these evolve independently but requires explicit interfaces and consistent propagation of execution and security context; separating system access from resource-specific enforcement supports least privilege, as the case's override rule illustrates. The case also shows that model-driven coordination need not transfer authority to the model. The model acts as an adaptive planner within an execution harness in which deterministic services, verifiers, and policy controls retain authority over consequential operations~\cite{lu2026safety}. Independent verification therefore constrains action selection without requiring the model itself to be trustworthy.

Separation has costs and limits. Physical separation adds interfaces, latency, and distributed failure modes; the worked example needs 27 exchanges for one question. Responsibilities should be physically separated where they require distinct credentials, trust, scaling, or change cadence, as L4 providers and L6 actions typically do. Otherwise they may be colocated, provided ownership and enforcement points remain distinct. Separation also does not provide confidentiality by itself: context sent to a hosted model, persistent memory, and telemetry are disclosure paths, so retrieval authorization (L5), provider data policy (L4), and retention (L9) need coordinated policy set at L0.

\subsection{Limitations}

The architecture rests on a targeted review and on researcher judgment in normalizing concepts. The instantiation is a research prototype of an enterprise integration scenario rather than a production deployment. It covers one domain and one existing system: semiconductor holdings for which emissions and revenue data are available, an illustrative fund mandate, and one hosted language model whose choices are not deterministic, with consistency reported over three executions per question. Because the instantiation was designed around the proposed boundaries, the probes show that the boundaries are enforceable and consistent, not that they are preferable to alternative decompositions. Section~\ref{sec:architecture-synthesis} argues that they reduce ambiguity of ownership, allow realizations to be substituted, and localize assurance, but comparative evaluation remains future work. The independent mapping interprets one source, and the evaluation does not address effects on quality attributes such as performance, maintainability, security, or cost.

\section{Conclusion}
\label{sec:conclusion}

We presented a technology-neutral reference architecture for enterprise agentic AI integration comprising ten responsibilities, four boundary invariants, eight typed interactions, and explicit variation points that separate stable responsibility ownership from condition-dependent capabilities and replaceable mechanisms. Instantiated as an ESG portfolio-analysis and pre-trade compliance agent over an existing ESG system, it realized seven responsibilities; different task and action conditions altered runtime interactions and activated conditional controls, consistent with the evaluated boundaries. An independently published agentic architecture was representable within L0--L9 without an additional responsibility. Future work will address further domains, multi-agent and asynchronous configurations, monitoring-driven runtime intervention, and effects on quality attributes.

\bibliographystyle{IEEEtran}
\bibliography{references}

@article{abouali2025agentic,
  author  = {Abou Ali, Mohamad and Dornaika, Fadi and Charafeddine, Jinan},
  title   = {Agentic {AI}: A Comprehensive Survey of Architectures,
             Applications, and Future Directions},
  journal = {Artificial Intelligence Review},
  volume  = {59},
  number  = {1},
  pages   = {11},
  year    = {2025},
  note    = {Article 11; published online 14 November 2025},
  doi     = {10.1007/s10462-025-11422-4}
}

@article{peykani2026holistic,
  author  = {Peykani, Pejman and Ghanidel, Sanly and Javadi-Sisi, Iman and
             Sn{\'a}{\v s}el, V{\'a}clav and Mirjalili, Seyedali},
  title   = {A Holistic Review of Agentic {AI} Frameworks, Applications,
             and Research Trajectories},
  journal = {Archives of Computational Methods in Engineering},
  year    = {2026},
  note    = {Published online 15 June 2026},
  doi     = {10.1007/s11831-026-10675-8}
}

@article{moralles2026slr,
  author  = {Moralles, Cassiano and da Costa, Luis Antonio L. F. and
             Rigo, Sandro Jos{\'e} and Kunst, Rafael and
             de Souza, Vin{\'i}cius Costa and Silva, {\'E}derson Passos and
             Prado, George Lucas Ebertz and
             Schardosim, Taimisson de Carvalho and Roehrs, Alex},
  title   = {A Systematic Literature Review of Agentic {AI}:
             Definitions, Architectures, and Challenges},
  journal = {IEEE Access},
  volume  = {14},
  pages   = {36176--36190},
  year    = {2026},
  doi     = {10.1109/ACCESS.2026.3668138}
}

@article{koo2026enterprise,
  author  = {Koo, Gi-Do and Chung, Myoungsug},
  title   = {Reference Architecture for Enterprise {AI} Transformation
             Using Agentic {AI}: A Design Science Approach},
  journal = {The Journal of Korean Institute of Communications and
             Information Sciences},
  volume  = {51},
  number  = {8},
  pages   = {1588--1609},
  year    = {2026},
  doi     = {10.7840/kics.2026.51.8.1588}
}

@inproceedings{lu2024responsible,
  author    = {Lu, Qinghua and Zhu, Liming and Xu, Xiwei and Xing, Zhenchang and
               Harrer, Stefan and Whittle, Jon},
  title     = {Towards Responsible Generative {AI}: A Reference Architecture
               for Designing Foundation Model Based Agents},
  booktitle = {2024 IEEE 21st International Conference on Software
               Architecture Companion (ICSA-C)},
  pages     = {119--126},
  year      = {2024},
  publisher = {IEEE},
  doi       = {10.1109/ICSA-C63560.2024.00028}
}

@inproceedings{elsisi2026devcoach,
  author    = {Elsisi, Omar and De Abreu Santos, F{\'a}bio and Melo, Glaucia},
  title     = {Lifecycle-Aware {GenAI} Assistance with {MCP} via Context
               Refinement Loops: A Reference Architecture},
  booktitle = {Proceedings of the 34th ACM International Conference on the
               Foundations of Software Engineering (FSE Companion '26)},
  pages     = {1614--1622},
  year      = {2026},
  publisher = {Association for Computing Machinery},
  doi       = {10.1145/3803437.3805617}
}

@inproceedings{galster2011empirically,
  author    = {Galster, Matthias and Avgeriou, Paris},
  title     = {Empirically-Grounded Reference Architectures: A Proposal},
  booktitle = {Proceedings of the Joint ACM SIGSOFT Conference on Quality of
               Software Architectures and ACM SIGSOFT Symposium on
               Architecting Critical Systems (QoSA-ISARCS)},
  publisher = {ACM},
  pages     = {153--158},
  year      = {2011},
  doi       = {10.1145/2000259.2000285}
}

@inproceedings{nakagawa2014consolidating,
  author    = {Nakagawa, Elisa Y. and Guessi, Milena and Maldonado, Jos{\'e} C.
               and Feitosa, Daniel and Oquendo, Flavio},
  title     = {Consolidating a Process for the Design, Representation, and
               Evaluation of Reference Architectures},
  booktitle = {2014 IEEE/IFIP Conference on Software Architecture (WICSA)},
  publisher = {IEEE},
  pages     = {143--152},
  year      = {2014},
  doi       = {10.1109/WICSA.2014.25}
}

@article{garousi2019guidelines,
  author  = {Garousi, Vahid and Felderer, Michael and M{\"a}ntyl{\"a}, Mika V.},
  title   = {Guidelines for Including Grey Literature and Conducting
             Multivocal Literature Reviews in Software Engineering},
  journal = {Information and Software Technology},
  volume  = {106},
  pages   = {101--121},
  year    = {2019},
  doi     = {10.1016/j.infsof.2018.09.006}
}

@article{galster2014variability,
  author  = {Galster, Matthias and Weyns, Danny and Tofan, Dan and
             Michalik, Bartosz and Avgeriou, Paris},
  title   = {Variability in Software Systems---A Systematic Literature Review},
  journal = {IEEE Transactions on Software Engineering},
  volume  = {40},
  number  = {3},
  pages   = {282--306},
  year    = {2014},
  doi     = {10.1109/TSE.2013.56}
}

@article{chand2026sra,
  author  = {Chand, Sailesh Saras and Kumar, Bimal Aklesh and
             Nakagawa, Elisa Yumi},
  title   = {Design and Evaluation of Software Reference Architecture:
             A Systematic Mapping Study},
  journal = {Applied Computing and Informatics},
  pages   = {1--17},
  year    = {2026},
  note    = {Ahead of print},
  doi     = {10.1108/ACI-05-2026-0344}
}

@misc{aws2026agentic,
  author       = {{Amazon Web Services}},
  title        = {Agentic {AI} Architecture in the Enterprise},
  howpublished = {AWS Prescriptive Guidance: Governing and Architecting the
                  Diversity of Agentic AI at Scale},
  year         = {2026},
  url          = {https://docs.aws.amazon.com/prescriptive-guidance/latest/govern-architect-agentic-ai/enterprise-architecture.html},
  note         = {Accessed: 2026-09-21}
}

@misc{microsoft2026multiagent,
  author       = {{Microsoft}},
  title        = {Multi-Agent Reference Architecture},
  howpublished = {GitHub Pages},
  year         = {2026},
  url          = {https://microsoft.github.io/multi-agent-reference-architecture/},
  note         = {Accessed: 2026-09-21}
}

@misc{google2026components,
  author       = {{Google Cloud}},
  title        = {Choose Your Agentic {AI} Architecture Components},
  howpublished = {Google Cloud Architecture Center},
  year         = {2026},
  url          = {https://docs.cloud.google.com/architecture/choose-agentic-ai-architecture-components},
  note         = {Last reviewed 21 April 2026. Accessed: 2026-09-21}
}

@misc{google2026multitenant,
  author       = {{Google Cloud}},
  title        = {Multi-Tenant Agentic {AI} System},
  howpublished = {Google Cloud Architecture Center},
  year         = {2026},
  url          = {https://docs.cloud.google.com/architecture/multi-tenant-agentic-ai-system},
  note         = {Last reviewed 18 June 2026. Accessed: 2026-09-21}
}

@techreport{mcp2026spec,
  author      = {{Model Context Protocol Project}},
  title       = {Model Context Protocol Specification, Version 2026-07-28},
  institution = {Agentic AI Foundation, The Linux Foundation},
  type        = {Specification},
  number      = {2026-07-28},
  year        = {2026},
  month       = jul,
  url         = {https://modelcontextprotocol.io/specification/2026-07-28},
  note        = {Originally developed by Anthropic; contributed to the Linux
                 Foundation's Agentic AI Foundation, December 2025.
                 Accessed: 2026-09-21}
}

@techreport{a2a2026spec,
  author      = {{A2A Project}},
  title       = {Agent2Agent ({A2A}) Protocol Specification, Version 1.0.1},
  institution = {The Linux Foundation},
  type        = {Specification},
  number      = {v1.0.1},
  year        = {2026},
  month       = may,
  url         = {https://a2a-protocol.org/v1.0.1/specification/},
  note        = {Originally developed by Google; hosted by the Linux
                 Foundation. Accessed: 2026-09-21}
}

@inproceedings{yu2026ontometric,
  author    = {Yu, Mingqin and Rabhi, Fethi and Suleiman, Basem and
               Liu, Nuodi and Li, Yuekang},
  title     = {An Ontology-Driven Service-Oriented System for {ESG} Metric
               Computation and Reporting},
  booktitle = {2026 IEEE International Conference on Web Services (ICWS)},
  pages     = {347--353},
  year      = {2026},
  month     = jul,
  publisher = {IEEE},
  doi       = {10.1109/ICWS72778.2026.00052}
}

@misc{lu2026safety,
  author       = {Lu, Qinghua and Bengio, Yoshua},
  title        = {{AI} Safety: Not Optional, Not Later},
  howpublished = {arXiv preprint arXiv:2609.10630},
  year         = {2026},
  url          = {https://arxiv.org/abs/2609.10630}
}

@techreport{tcfd2021implementing,
  author      = {{Task Force on Climate-related Financial Disclosures}},
  title       = {Implementing the Recommendations of the Task Force on
                 Climate-related Financial Disclosures},
  institution = {Financial Stability Board},
  year        = {2021},
  month       = oct,
  url         = {https://assets.bbhub.io/company/sites/60/2021/07/2021-TCFD-Implementing_Guidance.pdf},
  note        = {Defines weighted average carbon intensity for asset owners
                 and managers. Accessed: 2026-10-07}
}

\end{document}